\documentclass[letterpaper]{article}
\usepackage{aaai21}
\usepackage{times}
\usepackage{helvet}
\usepackage{courier}
\usepackage[hyphens]{url}
\usepackage{graphicx}
\usepackage{graphicx}
\usepackage{natbib}
\usepackage{caption}
\usepackage{xcolor}
\usepackage{booktabs}
\usepackage{amsmath,amsfonts,amssymb,amsthm,dsfont,stmaryrd}
\usepackage[switch]{lineno}

\newcommand\przemek[1]{{\color{blue} [\bf PS: #1]}}
\newcommand\igor[1]{{\color{magenta} [Ig: #1]}}

\newcommand\marcin[1]{{\color{red} [#1]}}

\def\D{\mathcal{D}}
\def\E{\mathcal{E}}

\def\G{\mathcal{G}}

\def\R{\mathbb{R}}

\def\X{\mathbf{X}}

\def\Z{\mathcal{Z}}

\def\our{LCW generator}
\def\trick{Latent Trick}
\def\cwsquare{CW$^2$}
\def\fashionmnist{Fashion-MNIST}
\def\celeba{CelebA}
\def\cwg{CW-Generator}
\def\swg{SW-Generator}

\begin{document}
%
\title{
Generative models with kernel distance in data space}
\author{
Szymon Knop, Marcin Mazur, Przemys{\l}aw Spurek, Jacek Tabor, Igor Podolak\\}

\affiliations{
Faculty of Mathematics and Computer Science\\
Jagiellonian University, Kraków, Poland\\
}
\maketitle

\begin{abstract}
Generative models dealing with modeling a~joint data distribution are generally either autoencoder or GAN based. Both have their pros and cons, generating blurry images or being unstable in training or prone to mode collapse phenomenon, respectively. The objective of this paper is to construct a~model situated between above architectures, one that does not inherit their main weaknesses. The proposed \our{} (Latent Cramer-Wold generator) resembles a~classical GAN in transforming Gaussian noise into data space. What is of utmost importance, instead of a~discriminator, \our{} uses kernel distance. No adversarial training is utilized, hence the name generator. It is trained in two phases. First, an autoencoder based architecture, using kernel measures, is built to model a~manifold of data. We propose a~\trick{} mapping a~Gaussian to latent in order to get the final model. This results in very competitive FID values.
\end{abstract}


\section{Introduction}
\emph{Generative modeling} is a~fast-growing area of machine learning which deals with modeling a~joint distribution of data. 
Generative modeling's key task is to train a~generator network to produce new samples from a~given data space by transforming samples from noise distribution. Training usually minimizes the dissimilarity between real and generated data distributions.

Widely used approaches to generative modeling are GANs and models based on autoencoder architecture. Autoencoder consists of an encoder
$\mathcal{E}\colon\mathcal{X}\longrightarrow{}\mathcal{Z}$
and a~decoder
$\mathcal{D}\colon\mathcal{Z}\longrightarrow{}\mathcal{X}$
(a generator) networks. Training boils down to minimizing a~tuned sum of a~reconstruction error and some measure of similarity between the distribution of encoded data  $P_{\mathcal{E}(X)}$ and a~given prior (noise) distribution $P_Z$ on the latent $\mathcal{Z}$.
GAN consists of a~generator  
$\mathcal{G}\colon \mathcal{Z}\longrightarrow \mathcal{X}$ and a~discriminator that distinguishes between samples from real
$P_{X}$ 
and ``fake'' $P_{\G(Z)}$
data distributions. It learns adversarially by utilizing a~\emph{minimax} game rule.

Both approaches have their pros and cons. Autoencoder based generative methods produce theoretically elegant generative models with the drawback that they tend to generate blurry samples when applied to natural images. On the other hand, their main advantage over GAN based models is that they allow fitting manifold of data and approximate probability distribution simultaneously~\cite{goodfellow2014generative}. Contrary to GANs, in autoencoder based models, each data point can easily be directly mapped into latent space, needed in conditional data generation. 

GAN based methods' main advantage is their ability to produce sharp images, almost indistinguishable from real ones. On the other hand, GANs are harder to train, unstable and may suffer from the “mode collapse” problem, where the resulting model is unable to capture all the true data distribution variability. GAN based models imitate real data well but frequently do not cover the training data set's entire space~\cite{heusel2017gans}.

The objective of this paper is to show that it is achievable to effectively train a~model that does not inherit the weaknesses of the above models, such as  blurry images or complex adversarial training, and provides for better results. The proposed \our{}\footnote{ The code is available \url{https://github.com/gmum/lcw-generator}} is obtained in a~two-stage training and uses kernel measures in all cost functions. The discriminator network is no longer necessary.

Our contributions can be summarized as follows:
\begin{itemize}
\item we introduce a~new \our{}, which resembles a~classical GAN in transforming Gaussian noise into data space and use kernel distance instead of adversarial training,
\item we show that kernel based metric can be used as reconstruction error in classical autoencoder based generative models by introducing a~\cwsquare{} model that achieves state-of-the-art FID scores in it's category,
\item we propose a~\trick{}, which can be applied to any model with latent space to construct density-based interpolations.
\end{itemize}


\begin{figure}
\begin{center} 
 \includegraphics[width=\columnwidth]{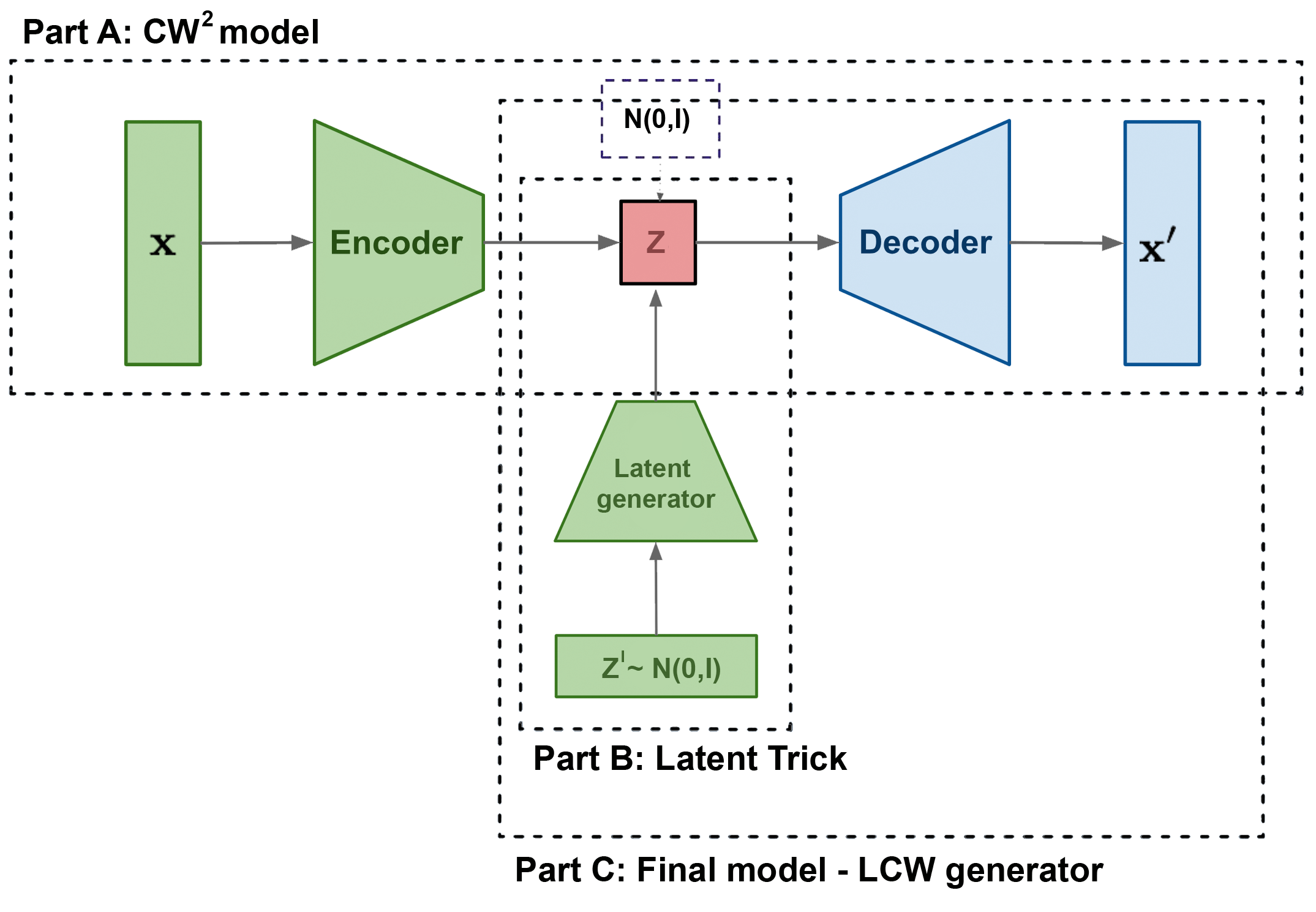}
\end{center} 
\caption{ The \our{} is trained in a~two-stage procedure. First, a~generative autoencoder (\cwsquare{} model) is trained (see Part~A). In the second stage, a~new Latent generator neural network is introduced and trained to transport a~standard Gaussian noise distribution $\mathcal{N}(0,I)$ into the autoencoder latent space (see Part~B). The final model is a~concatenation of the latent generator and the decoder (see Part~C).}
\label{fig:model} 
\end{figure} 

\section{Motivation}

In the case of simple data sets, e.g. MNIST or \fashionmnist{}, it is possible to effectively train a~generator network by directly minimizing an estimator of some distance between data and model distributions~\cite{deshpande2018generative,tabor2018cramer}. Unfortunately, such models may give poor results for the \celeba{} data set (see Tab.~\ref{tab_simple}).

At first glance, it seems that kernel distance  cannot be effectively applied in high dimensional spaces. But it turns out that the CW distance used in the CW autoencoder (CWAE) model introduced in~\citet{tabor2018cramer}, to measure dissimilarity between distribution of encoded data and the Gaussian prior on the latent space, can also be efficiently utilized as a~measure of reconstruction error (see the \cwsquare{} model introduction below). This autoencoder model obtains state-of-the-art results, suggesting that the problem does not lie in an objective function but in a~training procedure (see Tab.~\ref{tab_simple}).

\begin{figure}
\begin{center} 
\includegraphics[width=\columnwidth]{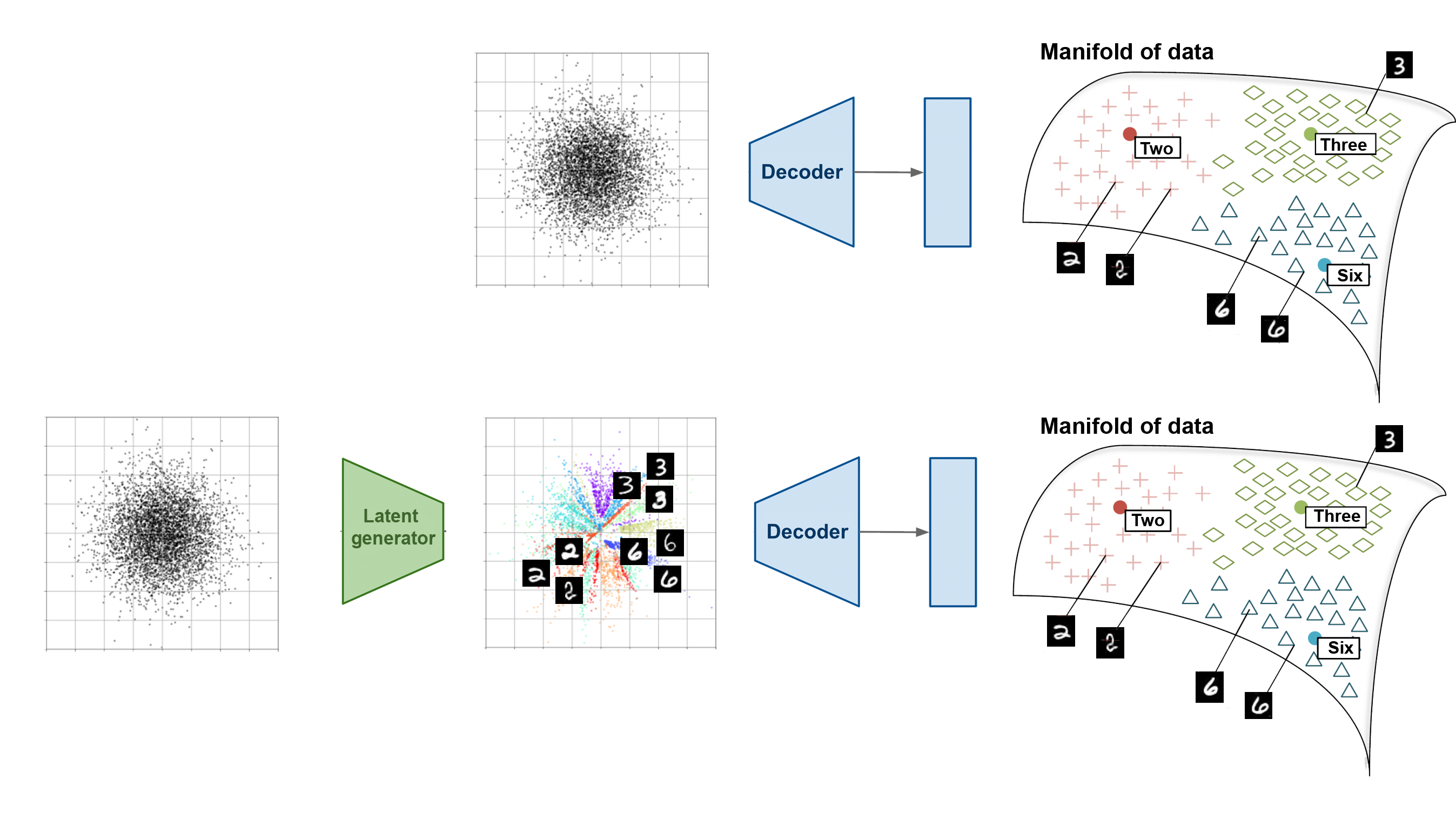}
\end{center} 
\caption{A classical decoder must simultaneously render manifold of data and transport a~given prior into data space to model data distribution (see top row). In the proposed modification, we first model manifold of data with autoencoder and then model probability distribution of data using a~Latent generator (see bottom row). This increases the generativity of the whole model. Visualization of data manifold inspired by~\cite{socher2013zero}.
}
\label{fig:ltrick} 
\end{figure}

Inferior results in previous to \cwsquare{} models were due to mini-batch training in high dimensional spaces with high noise. The model needs to minimize both the reconstruction error and the latent distribution distance from the prior, which might be complicated for non-simple data sets. In the proposed model, both are computed using kernel distances, and the training is two-step (see~Fig.~\ref{fig:ltrick}), which partly separates the responsibility for reconstruction/distribution training. Hence much better results are possible.



\section{General idea}\label{seq:cwg}

The proposed \our{} is trained in a~two-stage procedure (see Fig.~\ref{fig:model}). First, the \cwsquare{} model, a~modification of CWAE~\cite{tabor2018cramer}, is pretrained (see next section for details). Such architecture gives state-of-the-art FID score in class of generative autoencoder models but does not generate images as sharp as GANs do. 
To solve this problem the second stage of the construction is applied. There, the \trick{} (defined thoroughly in later section), in which the current autoencoder architecture is fixed and the new latent generator is trained to transport standard Gaussian noise distribution $\mathcal{N}(0,I)$ into the autoencoder latent space. This better approaches the encoded data distribution. The \trick{} is implemented as a~fully connected neural network. Consequently, the final model is given as concatenation of the latent generator and previously trained decoder. The \cwsquare{}~autoencoder may still be used on its own.

\begin{table}[htb]
\centering
\begin{tabular*}{\columnwidth}{@{\,}l@{\;\;\,}r@{\;\;\,}r@{\;\;\,}r@{\;\;\,}r@{\,}}
\toprule
        & CWAE & \cwg{} & \swg{} & \cwsquare{}\\
\midrule
MNIST   & 23.63 & \bf 16.48        & \bf 17.85 & \bf 17.44 \\ 
F-MNIST & 49.95 & \bf 31.05        & 48.02 & \bf 33.34\\ 
\celeba{} & \bf 49.69 & 87.01 & 141.03     & \bf 47.09\\
\bottomrule
\end{tabular*}
\caption{FID scores on MNIST, \fashionmnist{} and \celeba{} data sets obtained with CWAE (state-of-the-art autoencoder based generative model), classical \cwg{} (see~\citet{tabor2018cramer} for both), \swg{}~\cite{deshpande2018generative} and the proposed \cwsquare{}. 
}
\label{tab_simple}
\end{table}



We illustrate the concept of the model in Fig.~\ref{fig:ltrick}. Classical generative autoencoder's decoder (see the top row in Fig.~\ref{fig:ltrick}) must simultaneously render manifold of data and transport a~given prior distribution on the latent into data space to model data distribution. In the proposed solution, the data manifold is modeled first and then the probability distribution of data model is expanded using the Latent generator (see the bottom row in Fig.~\ref{fig:ltrick}).
Above is the crucial aspect of the proposed solution which inherits the positive properties of both autoencoder and GAN based generative methods. A~stable model is obtained, with precise autoencoder latent space producing high-quality images without adversarial training.

\begin{figure*}[htb]
\centering
\begin{tabular}{@{}c@{}c@{}c@{}}
 \rotatebox{90}{ \qquad\qquad CWAE+LT} & \,
\includegraphics[height=4.8cm]{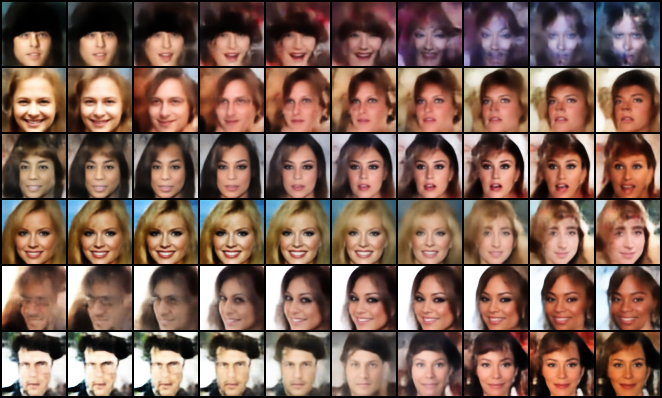}& \,
\includegraphics[height=4.8cm]{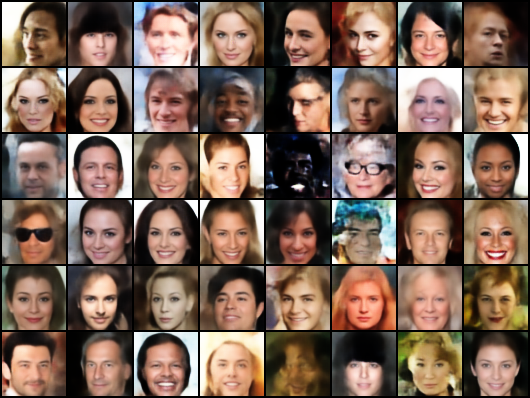} \\
 \rotatebox{90}{ \qquad\qquad LCW} & \,
\includegraphics[height=4.8cm]{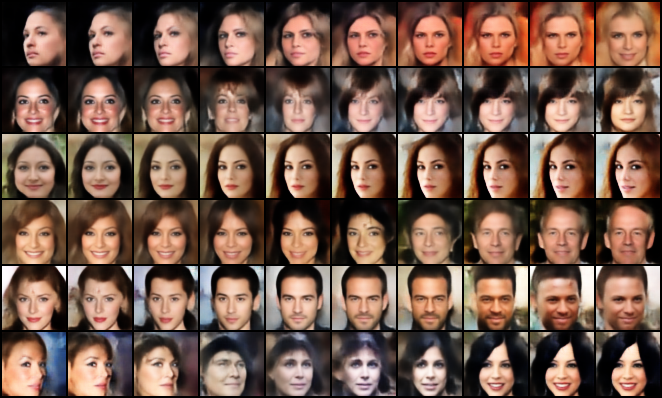}& \,
\includegraphics[height=4.8cm]{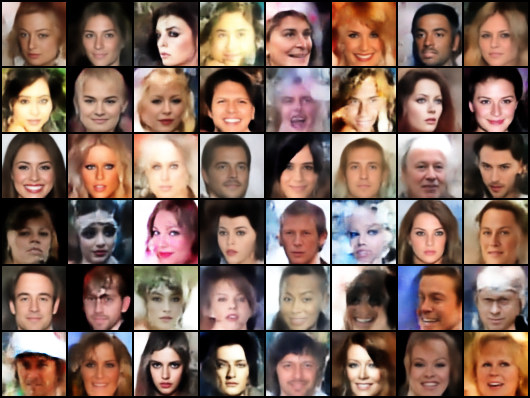} 
\end{tabular}
\caption{Results of CWAE+LT and \our{} models trained on \celeba{} data set.  As we can see, we obtain state-of-the-art samples and interpolations in autoencoder based generative models.}
\label{fig:celeb}
\end{figure*}

\section{CWAE and \cwsquare{} models}\label{seq:cw2}

In general, all autoencoder based generative models are trained to minimize an objective function of the form
\begin{linenomath*}
\begin{equation*}
\begin{array}{c}
\mathcal{J}(X;\E,\D)=\mathrm{Err}(\X,\D(\E (X) ))  +  \lambda \, \mathrm{DM}(P_{\E (X)},P_Z),
\end{array}
\end{equation*}
\end{linenomath*}
where $\mathrm{Err}$ is a~reconstruction error term, $\lambda$ is a~hyperparameter and DM 
denotes any dissimilarity, not necessarily non-negative, measure between probability distributions on the latent $\Z$.
CWAE uses the mean squared error MSE
, logarithm of the square of the CW distance $d_\mathrm{CW}$ and $\mathcal{N}(0,I)$ as the prior $P_Z$, which leads to the following formula
\begin{linenomath*}
\begin{equation*}
\begin{array}{c}
\mathrm{MSE}(X,\D(\E (X) ))  +  \lambda \log{d^2_\mathrm{CW}}(P_{\E (X)},\mathcal{N}(0,I)).
\end{array}
\end{equation*}
\end{linenomath*}
Following~\citet{tabor2018cramer}, we emphasize that $d_\mathrm{CW}$ can be calculated analytically and approximated as a~distance between a~latent sample $Z=(z_i)_{i=1}^n$ and the standard Gaussian prior. Specifically,
\begin{linenomath*}
\begin{equation*} 
\begin{array}{c}
2\sqrt{\pi}\,d^2_\mathrm{CW}(Z,\mathcal{N}(0,I)) \, \approx \, \tfrac{1}{n^2}\sum_{ij} 
(\gamma_n+\tfrac{\|z_i-z_j\|^2}{2D_\mathcal{Z}-3})^{-\frac{1}{2}}\\ +\,(1+\gamma_n)^{-\frac{1}{2}} - \tfrac{2}{n} \sum_{i} (\gamma_n+\tfrac{1}{2}+\tfrac{\|z_i\|^2}{2D_\mathcal{Z}-3})^{-\frac{1}{2}},
\end{array}
\end{equation*}
\end{linenomath*}
consequently giving the following CWAE cost function
\begin{linenomath*}
\begin{equation*}
\begin{array}{c}
\mathcal{J}_\textrm{CW}= \mathrm{MSE}(X,\D(\E (X) ))  +  \lambda \log{d^2_\mathrm{CW}}(\E (X),\mathcal{N}(0,I)),
\end{array}
\end{equation*}
\end{linenomath*}
where $D_\mathcal{Z}=\mathrm{dim}\,\mathcal{Z}$ and $\gamma_n$ bandwidth is chosen using Silverman's rule of thumb, i.e. $\gamma_n=\hat{\sigma}(\frac{4}{3n})^{2/5}$, where  $\hat{\sigma}$ denotes a~standard deviation that we assume to be equal to 1 as we deal with the standard Gaussian distribution.

We want to point out that the CW distance is given by a~characteristic kernel with a~closed-form for spherical Gaussians. Moreover, to the best of our knowledge, CWAE model was the first kernel distance based concept that required no sampling from the prior distribution.

Taking into consideration the results of~\citet{tabor2018cramer}, it is also possible to obtain the following approximate analytical formula that expresses $d_\mathrm{CW}$ for two given samples $X=(x_i)_{i=1}^n$ and $Y=(y_i)_{i=1}^n$in $\mathcal{X}$
\begin{linenomath*}
\begin{equation*} 
\begin{array}{c}
2\sqrt{\pi}\,d^2_\mathrm{CW}(X,Y) \;\, \approx \;\,
\tfrac{1}{n^2}\sum_{ij} 
(\gamma_n+\tfrac{\|x_i-x_j\|^2}{2D_\mathcal{X}-3})^{-\frac{1}{2}} \,+ \\ \tfrac{1}{n^2}\sum_{ij} 
(\gamma_n+\tfrac{\|y_i-y_j\|^2}{2D_\mathcal{X}-3})^{-\frac{1}{2}}- \tfrac{2}{n} \sum_{ij} (\gamma_n+\tfrac{\|x_i-y_j\|^2}{2D_\mathcal{X}-3})^{-\frac{1}{2}},
\end{array}
\end{equation*}
\end{linenomath*}
where $D_\mathcal{X}=\mathrm{dim}\,\mathcal{X}$ and $\gamma_n$ is calculated using standard deviation of joined $X$ and $Y$ samples.

This suggests using the CW~distance not only in the latent, but also in the data space as $\mathrm{Err}(\X,\D(\E (X) ))$. Consequently we introduce the \cwsquare{} model with the following objective function\footnote{It should be mentioned here that this is not a~classical autoencoder function, because a~reconstruction error does not depend directly on differences between $x_i$ and $\D(\E(x_i))$ as in MSE
, but is based on a~distance between distributions.}
\begin{linenomath*}
\begin{equation*}
\begin{array}{c}
\mathcal{J}_{\mathrm{CW}^2} = d^2_\mathrm{CW}(X,\D(\E (X) ))  +  \lambda \log{d^2_\mathrm{CW}}(\E (X),\mathcal{N}(0,I)),
\end{array}
\end{equation*}
\end{linenomath*}
minimized in \our{} construction's first phase.
The square sign in \cwsquare{} denotes that it uses the CW distance twice.

\section{\trick{}}\label{lt}

\begin{figure}[htb]
\normalsize
\begin{center}
\begin{tabular}{@{}c@{}c@{}c@{}}
 AE & CWAE & \cwsquare{} \\
\includegraphics[width=0.15\textwidth]{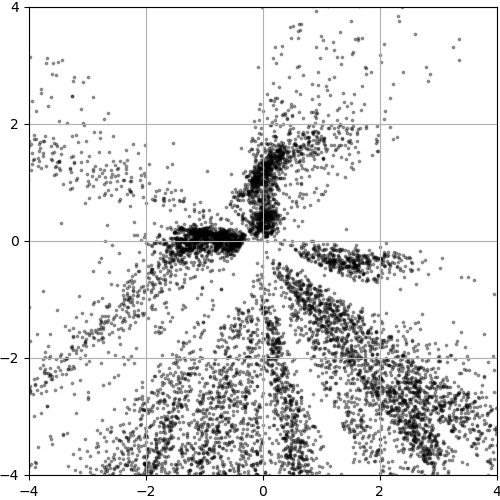}&
\includegraphics[width=0.15\textwidth]{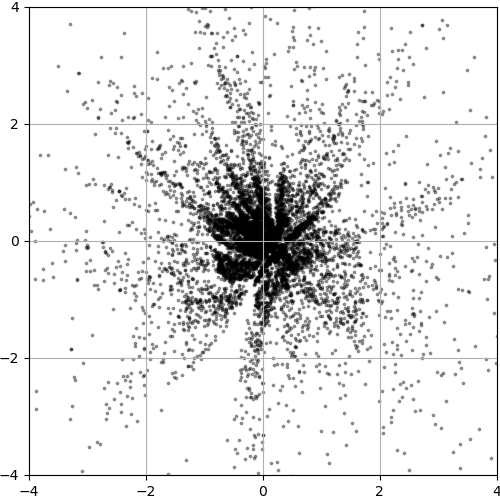}&
\includegraphics[width=0.15\textwidth]{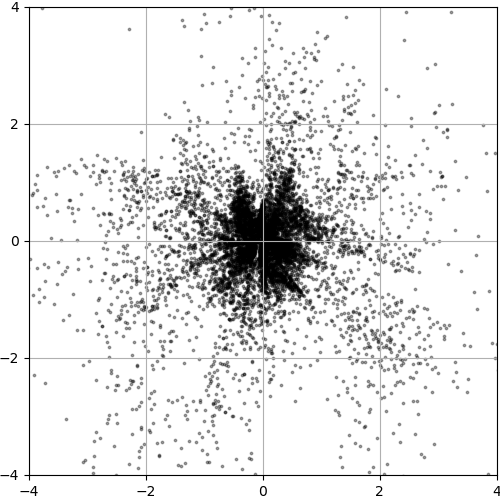} \\
 AE+LT & CWAE+LT & LCW \\
\includegraphics[width=0.15\textwidth]{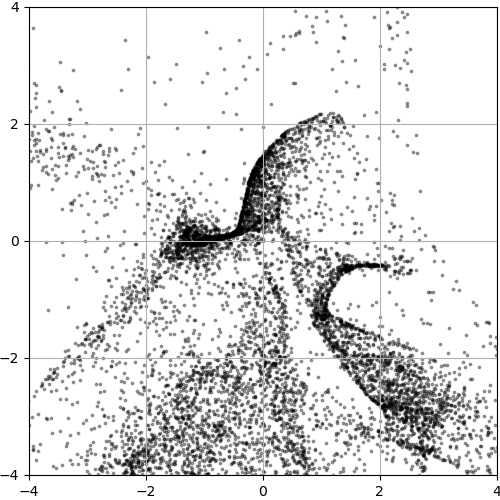}&
\includegraphics[width=0.15\textwidth]{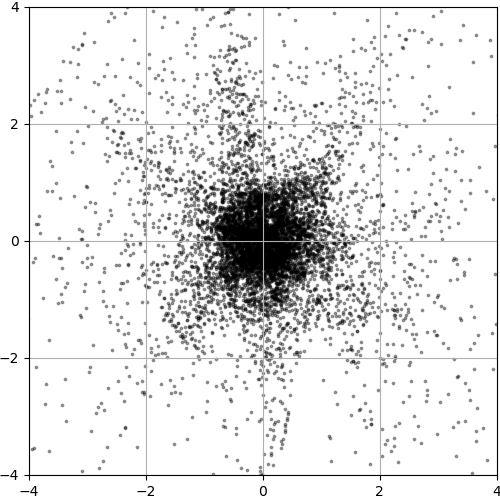}&
\includegraphics[width=0.15\textwidth]{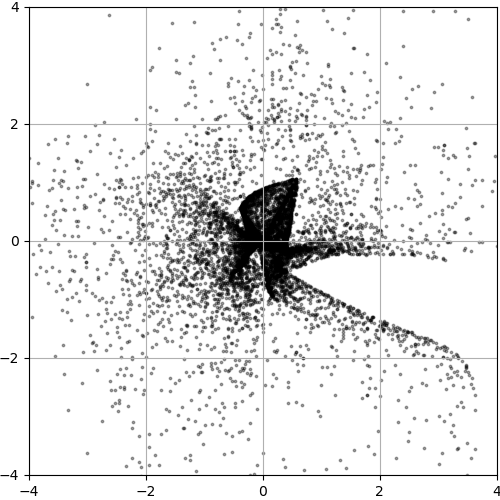}
\end{tabular}
\end{center}
\caption{The top row shows encoded data distributions learned by AE, CWAE and CW$^2$ models on \fashionmnist{}'s validation data set. Note that they are close to but different from the standard Gaussian noise. However, they are very similar to bottom row figures, which present the \our{}'s output, applied for the above models, i.e the standard Gaussian distribution transported to respective latent spaces.}
\label{fig:latent2D} 
\end{figure}

Note that in the first stage (see Part~A in Fig.~\ref{fig:model}) of model construction, the latent distribution is forced to be as close as possible to the Gaussian prior. Hence the obtained \cwsquare{} model is certainly generative.

\begin{table}[htb]
\normalsize
\begin{center}
{
\begin{tabular*}{\columnwidth}{@{\,}l@{\,\,}l@{\,}r@{\,\,}r@{\,\,}r@{\,\,}r@{\,\,}r@{\,\,}r@{\,\,}r@{\,\,}r@{\,\,}r@{\,}}
\toprule
Data set & Method  & Learn. & $\lambda$ & Latent & Noise & Rec. & FID \\
 & & rate & & dim & dim & error & score\\
\midrule             
MNIST   & AE & 1.e-3 & - & 8 & - & 11 & 52 \\
        & AE+LT & 5.e-4 & - & 8 & 8 & - & 22 \\
        & CWAE & 1.e-3 & 1 & 8 & - &11 & 23 \\
        & CWAE+LT & 5.e-4 & - & 8 & 8 & - &  20\\
        & \cwsquare{} & 1.e-3 & 1 &  8 & - & 14 & \bf 17  \\
        & LCW & 5.e-4 & - &  8 & 8 & - &  \bf 14  \\
\midrule        
F-MNIST & AE+LT & 5.e-4 & - &  8 & 8 & - & 41 \\
        & CWAE & 1.e-3 & 10 &  8 & - & 10 & 49 \\
        & CWAE+LT & 5.e-4 & - & 8 & 8 &   - & 38  \\
        & \cwsquare{} & 1.e-3 & 1 &  8 & - & 13 & \bf  33 \\
        & LCW & 5.e-4 & - &  8 & 8 &  - &  \bf 28  \\
\midrule        
\celeba{} & AE & 1.e-3 & - & 64 & - & 66 &   328  \\
        & AE+LT & 5.e-4 & - &  128 & 32 &  - & 45  \\
        & CWAE & 5.e-4 & 5 &  64 & - &  68  & 49  \\
        & CWAE+LT & 5.e-4 & - &  128 & 32 & - & \bf 31 \\
        & \cwsquare{} & 5.e-4 & 0.2 &  64 & - & 71  &  47  \\
        & LCW & 5.e-4 & - & 128 &  32 & -  & \bf 33  \\
\bottomrule
\end{tabular*}
}
\end{center}
\caption{Comparison of different architectures on MNIST, \fashionmnist{} and \celeba{} data sets. All models outputs except AE are similarly close to the normal distribution. CWAE achieves the best value of FID score (lower is better). 
All hyperparameters were found using a~grid search (see appendix).}
\label{tab:ablation}
\end{table}

However, there remain two fundamental problems. One is because there are empty holes/spaces in the latent space , i.e. parts with very low data density being mapped. In the classical autoencoder based approach, there is a~need for a~compromise between reconstructing and generating abilities. A~possible solution is to use an appropriate hyperparameter to balance the two terms~\cite{higgins2017beta} but, in practice, it increases the generativity of the model at the expense of reconstruction.

Second problem seems to be more fundamental~\cite{li2015generative,dziugaite2015training,li2017mmd}. It is related to
using mini-batch training in high dimensional data space with much lower intrinsic data dimension. Each batch contains only a~small, typically unbalanced, subset of a~data set. We hypothesize that it affects kernel methods' effectiveness because they need to learn representation and data distribution simultaneously, see Fig.~\ref{fig:ltrick}.


To solve the above problems, we introduce the \trick{}, which is the second stage of our procedure. It involves the creation of a~latent generator 
\begin{linenomath*}
\begin{equation*}
\mathcal{LG}\colon (\mathcal{Z}',\mathcal{N}(0,I))\longrightarrow{} (\mathcal{Z},P_{\E(X)}),
\end{equation*}
\end{linenomath*}
which is a~neural network trained to transform the standard Gaussian distribution (on the new $\mathcal{Z}'$ space) so that it resembles the distribution of encoded data on \cwsquare{} model's latent~$\mathcal{Z}$. To be precise, the objective is to minimize the following function
\begin{linenomath*}
\begin{equation*}
\mathcal{J}_\mathrm{LT}(X,Z';\mathcal{LG})=d^2_\mathrm{CW}(\E(X),\mathcal{LG}(Z'))),
\end{equation*}
\end{linenomath*}
where $X$ and $Z'$ denote a~data sample and a~sample from $\mathcal{N}(0,I)$ distribution, respectively. Consistently, to express the CW distance an appropriate approximation formula is used (analogous to that in the previous section).

Note that the \trick{} phase allows to rectify model's generative power without losing it's reconstruction quality. Consequently, the final \our{} is provided as a~function
$\mathcal{LG}_\mathrm{CW} \colon \mathcal{Z'}\longrightarrow{}\mathcal{X}$, which transport a~Gaussian noise sample $Z'$ into the data space, via concatenation of $\mathcal{LG}$ and $\D$ (compare diagram in Fig.~\ref{fig:model}), i.e.
\begin{linenomath*}
\begin{equation*}
\mathcal{LG}_\mathrm{CW}(Z')=\D(\mathcal{LG}(Z')).
\end{equation*}
\end{linenomath*}

 The influence of the \trick{} is visualized in Fig~\ref{fig:latent2D}. Top row shows the mapping of the whole data set on the latent (in $\mathbb{R}^2$ here for simplicity) for AE, CWAE, and \cwsquare{} models are shown, while the bottom row shows mapping of the Gaussian noise through \trick{} onto the same space. The corresponding figures seem to differ slightly, i.e. the $\mathcal{LG}$ mappings can be understood as generalizations of the encoded data samples. Hence the small discrepancies.

\trick{} can be used not only for the above construction but to any classical autoencoder based generative model that uses MSE as a~reconstruction error. We examine some of these models in the Experiments section.

\begin{figure*}[htb]
\centering
\begin{tabular}{@{}c@{}c@{}c@{}}
AE &  CWAE &  CW${^2}$ \\
\includegraphics[height=5.2cm]{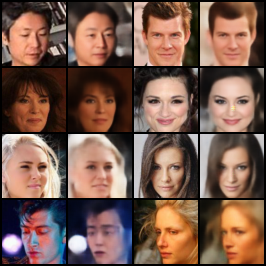}& \,
\includegraphics[height=5.2cm]{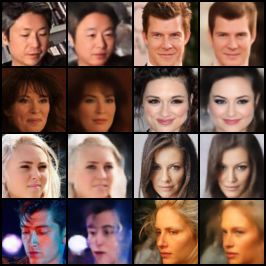} & \,
\includegraphics[height=5.2cm]{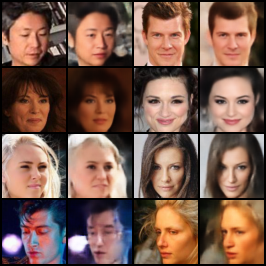}  \,
\end{tabular}
\caption{Reconstructions quality of AE, CWAE and \cwsquare{} (first stage of \our{}) models trained on \celeba{} data set. Odd columns correspond to the real test points, while even to their reconstructions.}
\label{fig:celeb_rec}
\end{figure*}

\section{Related work}

We divided the related work section into two parts. First, we describe existing approaches to train GAN style models (Generators) with kernel measures in data space. Then we discuss existing solutions improving autoencoder properties by adding a~neural network in latent space.

\paragraph{Generators}

Generative Moment Matching Network (GMMN)~\cite{li2015generative,dziugaite2015training} is a~deep generative model that differs from Generative Adversarial Network (GAN)~\cite{goodfellow2014generative} in replacing  GAN's discriminator with a~two-sample test based on kernel maximum mean discrepancy (MMD)~\cite{gretton2012kernel}. Unfortunately, these models work only for reasonable simple data sets like MNIST and \fashionmnist{}.

To solve such problems~\cite{li2017mmd} propose an MMD GAN. Authors improve the model expressiveness of GMMN and its computational efficiency by introducing adversarial kernel learning to replace a~fixed Gaussian kernel in the original GMMN. The proposed algorithm is similar to GAN, aiming to optimize two neural networks in a~minimax setting, but the objective's meaning is different. In GAN we train a~discriminator (binary) classifier to distinguish two distributions. MMD-GAN discriminates two distributions with a~two-sample test via MMD, but with an adversarially learned kernel. 
It is similar to our approach in using kernel distance but, contrary to \our{} proposed, still uses adversarial training.

\citet{deshpande2018generative} show that it is possible to train a~GAN like architecture (namely an \swg{})  by substituting discriminator with kernel measure in data space. In practice, the authors use a~Sliced Wasserstein distance. Since the sliced method can reduce data dimensionality by random projections, a~model can be trained effectively using kernel measures in high dimensional spaces. Unfortunately, their model works only for reasonable simple data sets like MNIST and \fashionmnist{}.

\paragraph{Autoencoder based generative models}
First and one of the most popular approaches to autoencoder based generative models is  VAE~\cite{kingma2014auto}. In practice, such models give correct results, but geometry of latent space might have some very low probability areas (empty spaces)~\cite{tolstikhin2018wasserstein}. An idea to join the training algorithms and, following it, the merits of autoencoders and GANs, was presented early in a~widely cited paper~\cite{larsen2016autoencoding}. Another method to solve such problems can be obtained by adding additional architecture to the latent, in order  to obtain better representation. For example, 
\citet{ziegler2019latent,xiao2019generative} apply  normalizing flows~\cite{kingma2018glow} in the latent space. Thanks to this the latent distribution (which is similar to the standard Gaussian) is transformed into the Gaussian prior. \citet{dai2018diagnosing}, on the other hand, present similar solution based on adding additional autoencoder in the latent space (TwoStageVAE model). They present theoretical results stating that VAE model does not properly approximate properly prior distribution in the latent space. But, as they propose, the second VAE model is able to correct distribution in the latent. \citet{deja2020end} use Sinkhorn autoencoder with Noise Generator (e2e SAE), a~simple fully connected architecture which transfers Gaussian noise into the latent space. This model is trained end-to-end and gives similar results to WAE-MMD~\cite{tolstikhin2018wasserstein}.

\section{Experiments}

In this section, we empirically validate the proposed \our{} model on standard benchmarks for autoencoder based generative models: \celeba{}, MNIST and \fashionmnist{}. Since MNIST and \fashionmnist{} are relatively simple, we use them rather as toy examples. \celeba{} data sets will be used to show real difference between models, see Tab.~\ref{tab:my_label2}.

It should be mentioned that GAN models obtain essentially better results. But our goal is not to outperform GANs but rather to reduce the~gap between the generative quality of the GAN based and non-adversarial AE based models. 
At the same time, as a~``by-product'', we propose the  \cwsquare{} model. Still being an autoencoder based model, it obtains much better FID values than other autoencoders (see, e.g., Tab.~\ref{tab:my_label1}).

\begin{table}[htb]
\centering
\begin{tabular*}{\columnwidth}{@{\,}l@{\;\;}r@{\;\,}r@{\;\,}r@{\,}}
\toprule
        & \cwg{} & \swg{} & \our{} \\
\midrule
MNIST   &   16.48        &  17.85 & \bf 14.92 \\ 
F-MNIST &   31.05        & 48.02 & \bf 28.04\\ 
\celeba{} &  87.01 & 141.03     & \bf 33.07 \\
\bottomrule
\end{tabular*}
\caption{FID scores on MNIST, \fashionmnist{} and \celeba{} data sets obtained with \cwg{}, \swg{} and \our{}. 
}
\label{tab:my_label2}
\end{table}


\begin{table}[htb]
\centering
\begin{tabular}{@{\;}l@{\;\;}r@{\;\;}l@{\;\;}r@{\;\;}l@{\;\;}r@{\;}}
\toprule
Model&FID&Model&FID&Model&FID\\
\midrule
VAE  & 60 & \bf 2Stage-VAE$^*$  & \bf 34  & \bf CWAE + LT  &  \bf 31\\
CWAE & 49 &CW$^2$  & 47  & \bf LCW & \bf 33\\
WAE  & 52 &WAE-GAN$^*$ & 42  & e2e SAE$^*$ & 54\\
\midrule
DCGAN$^*$ & 12.5& WGAN-GP$^*$ & 20.3 & WGAN-div$^*$& 17.5\\
\bottomrule
\end{tabular}
\caption{FID values for several compared models, computed on the \celeba{} data set. The \trick{} addition expands the model generative capabilities.  Models marked with an asterisk have results taken from literature. The DCGAN value was trained with a special two time scale algorithm~\cite{heusel2017gans}. WGAN have convolutional architectures and the results are taken from~\cite{Wu2017wasserstein}.}
\label{tab:my_label1}
\end{table}

\begin{figure*}[htb]
\centering
\begin{tabular}{@{}c@{\,}c@{\,}c@{}}
 &  Test interpolation &  Random sample \\
 \rotatebox{90}{ \qquad\qquad AE} & 
\includegraphics[height=4.8cm]{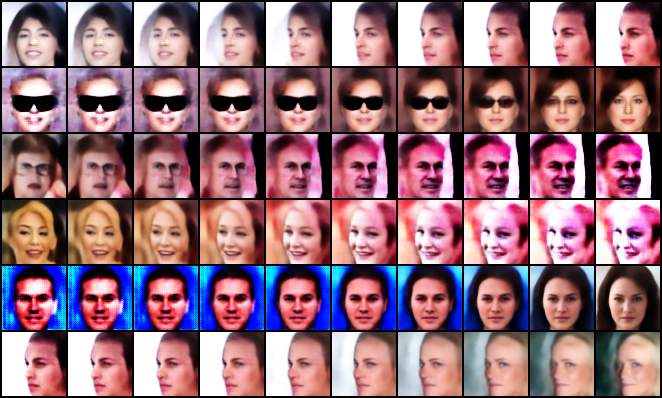}& 
\includegraphics[height=4.8cm]{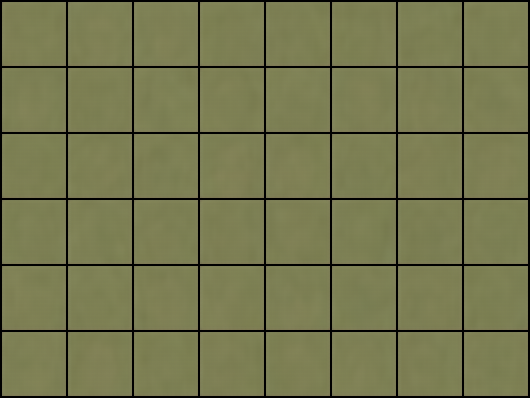}  \\
 \rotatebox{90}{ \qquad\qquad AE+LT} & 
\includegraphics[height=4.8cm]{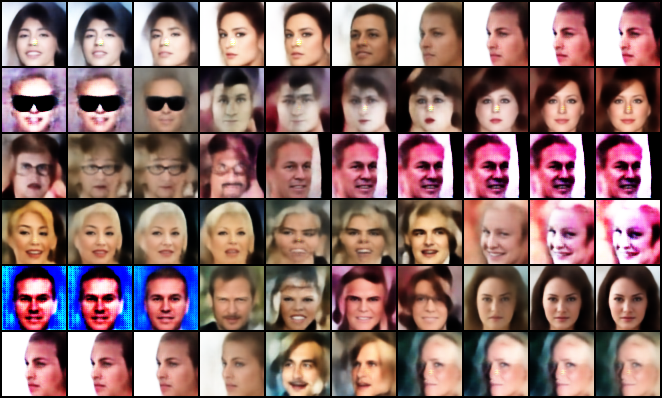}& 
\includegraphics[height=4.8cm]{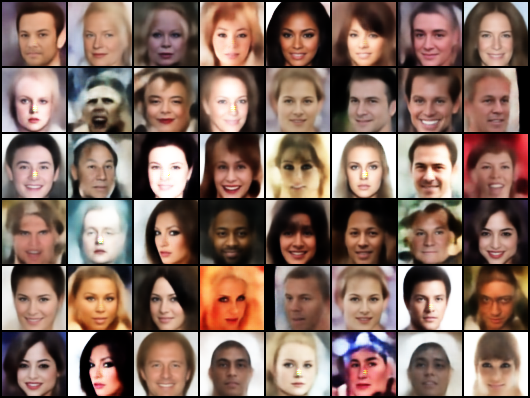} 

\end{tabular}
\caption{Results of AE and AE+LT models trained on \celeba{} data set. Interpolation in AE (top row) are constructed linearly in the  latent between ``endpoint'' images from AE+LT, while in AE+LT a linear interpolation between samples from $\mathcal{N}(0,I)$ is performed to be mapped as $\mathcal{LG}(z')$ points. This corresponds to two types of interpolations in Fig.~\ref{fig:laten_inter}. As we can see, classical AE is not a~generative model (top right). By applying \trick{} we produce  generative model AE+LT (bottom right).}
\label{fig:celeb_aelt}
\end{figure*}

\paragraph{Quantitative tests}
To quantitatively compare \our{} with  other models, in the first experiment we compare different models which use kernel distance in data space, i.e. CW--, SW-- and \our{}s. To assess results the Fr\'{e}chet Inception Distance FID is used~\cite{heusel2017gans}. As it can be easily seen, \our{} outperform other approaches. In the case of MNIST and \fashionmnist{} data set the difference is smaller than in the case of \celeba{}, since the two first data sets are quite simple.

\begin{figure}[htb]
\begin{tabular*}{\columnwidth}{@{}c@{}c@{}c@{}}
\includegraphics[width=0.16\textwidth]{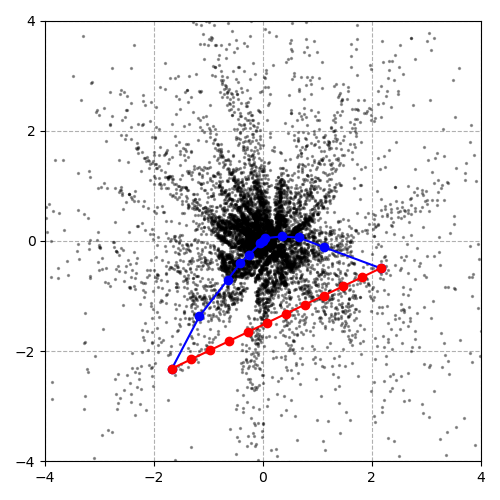}& 
\includegraphics[width=0.16\textwidth]{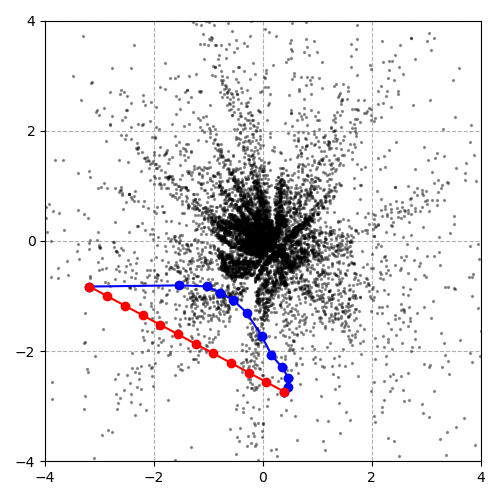}&
\includegraphics[width=0.16\textwidth]{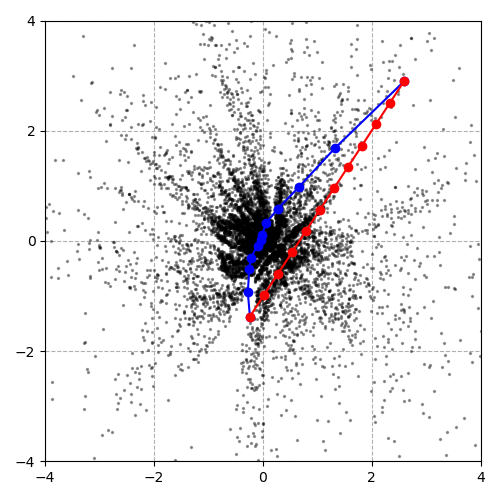}\\
\includegraphics[width=0.16\textwidth]{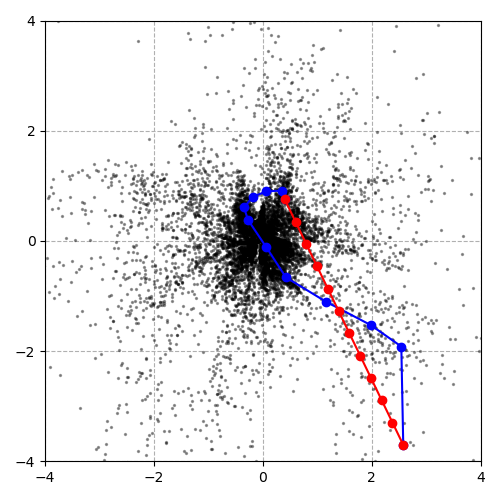}& 
\includegraphics[width=0.16\textwidth]{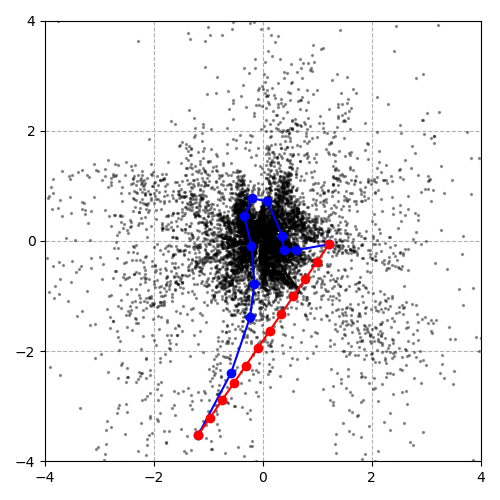}& 
\includegraphics[width=0.16\textwidth]{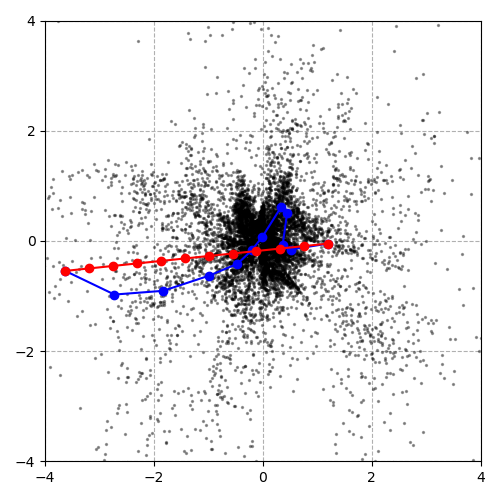}
\end{tabular*}
\caption{Interpolations in CWAE+LT (top row) and LCW generator (bottom row) models' latent spaces, trained on \celeba{} data set. The red dots show a GAN like linear interpolation $\alpha_i\mathcal{LG}(z'_1)+(1-\alpha_i)\mathcal{LG}(z'_k)$ for $z'_1,z'_k\sim\mathcal{N}(0,I)$, while the blue ones are obtained by mapping a~linear interpolation 
$\alpha_i{}z'_1+(1-\alpha_i)z'_k$ via $\mathcal{LG}$. In the latter case curves follow the latent distribution and avoid empty spaces.}
\label{fig:laten_inter}
\end{figure}

In the other experiment we follow the classical competition setup for generative models on \fashionmnist{} data set. In agreement with the qualitative studies, we observe FID values of \our{} to be better (lower) than those for VAE, CWAE, SWAE \cite{kolouri2018sliced}, WAE-MMD and WAE-GAN \cite{tolstikhin2018wasserstein}.

We stress here that \our{} on \celeba{} achieves FID score 33, compared to 52 and 42 achieved by WAE-MMD and WAE-GAN (see Tab.~\ref{tab:my_label1}). It should be mentioned that \our{} gives similar results to 2Stage-VAE and CWAE+LT, which also use two stage training but in first part use MSE as a~reconstruction cost.

\paragraph{Ablation study}
In this paragraph, we show how the proposed two-stage training procedure works in various autoencoder-based models. As it was already mentioned, the \trick{}
is a~possible solution to the problem with simultaneous modeling of a~manifold of data and generativity of the model. We claim that we model the manifold of data in the first stage, while the latent distribution in the second stage .

To test whether this hypothesis is true, we apply the \trick{} to some models. We start with a~standard autoencoder (AE), which models only the manifold of data. It turns out that such models work surprisingly well (see the difference between AE and AE+LT methods in Tab.~\ref{tab:ablation}). It should be mentioned that AE+LT gets FID score of 45 on \celeba{}, which is comparable to that of WAE-GAN's (42) and WAE-MMD's (52). It is still an open question why the \trick{} added to CWAE and \cwsquare{} models performs essentially better score than AE+LT (see Tab.~\ref{tab:ablation}). It seems that in AE we are able to model data manifold but latent representation is not constrained, so the \trick{} is not able to describe distribution of data efficiently. In~ Fig.~\ref{fig:celeb_aelt} we present qualitative tests obtained by interpolation and sampling.

Summarizing, \trick{} can be used to transform any autoencoder based architecture into generative model.  

\paragraph{Qualitative tests}
Since the \our{}, similarly to GAN models, does not produce reconstructions, we compare only those of \cwsquare{} model (trained as the first stage of \our{}) to ones produced by CWAE and vanilla AE, see Fig.~\ref{fig:celeb_rec}. We want to stress that \cwsquare{} produces quite accurate reconstructions even though it is not explicitly required (point to point) by its cost function.



\paragraph{Interpolations}

It is challenging to see the different interpolation abilities of the proposed \our{} model and its parts. The least interesting is that using the \cwsquare{} model only since it is a~simple, autoencoder like, interpolation in the latent $\mathcal{Z}$ space. Thus we shall skip it.

On the other hand, the construction of the \trick{} allows to make two interesting interpolations. First, let us draw $z'_1, z'_k\sim\mathcal{N}(0,I)$, i.e. two points from \trick{}'s input space $\mathcal{Z}'$. Mapping $z_1=\mathcal{LG}(z'_1), z_k=\mathcal{LG}(z'_k)$ it is possible to perform a~GAN like type linear interpolation $z_1,z_2,\dots,z_k$ 
in the latent $\mathcal{Z}$. On the other hand, there is a~possibility to produce a linear interpolation $z'_1,z'_2,\dots,z'_k$ 
in $\mathcal{Z}'$ and map all those points  to $\mathcal{LG}(z'_1),\mathcal{LG}(z'_2), \dots,\mathcal{LG}(z'_k)$. This produces a~density-based interpolation since $\mathcal{LG}$ network is trained to generate the latent distribution. The results can be seen in Fig.~\ref{fig:laten_inter}, where red dots show the standard like interpolation, while blue ones are obtained using the \trick{}. Note that in the latter case the model can follow areas of high data density.

Additionally, using all considered models we can construct interpolation and sampling, see Fig.~\ref{fig:celeb}. In the case of sampling we can see that \our{} gives state-of-the-art autoencoder based faces.

The above clearly shows that the proposed \our{} model can better map the latent space. Several methods for interpolation in latent spaces provide sophisticated approaches to computing the latent space data mapping density, frequently using Riemannian curvatures, see e.g.~\cite{Agustsson2017, Arvanitidis2017,LesniakSieradzkiPodolak:2019}. The proposed model generates density-based interpolations thanks only to applying the \trick{}.

\subsection{Architecture}\label{app:architectures}

For CWAE model we used hyperparameters reported in~\citet{tabor2018cramer}. For \cwsquare{} model, \swg{} and \cwg{}  we performed a~grid search over batch-size in $\{64, 128, 256, 512\}$ and learning rate values in $\{0.005, 0.0025, 0.001, 0.0005, 0.00025\}$. For every model, we reported results for configuration that achieved the lowest value of FID Score.

Autoencoder feed-forward architecture for MNIST or \fashionmnist{} ($28\times28$ images)
\begin{itemize}
\item[] \textbf{encoder} three feed-forward ReLU layers, 200 neurons each,
\item[]\textbf{decoder} three feed-forward ReLU layers, 200 neurons each followed by feed-forward sigmoid layer.
\end{itemize}

Autoencoder  architecture for \celeba{} (with images centered and cropped to $64\times64$ with 3 color layers): 
\begin{itemize}
\item[]\textbf{encoder}
\begin{itemize}
\item[] four convolution layers with $4\times4$ filters, each layer was followed by a~batch normalization (consecutively 128, 256, 512, and 1024 channels) and ReLU activation,
\end{itemize}
\item[]\textbf{decoder} 
\begin{itemize}
\item[] dense 1024 neuron layer,
\item[] three transposed-convolution layers with $4\times4$ filters, and each layer followed by a~batch normalization with ReLU activation (consecutively 512, 256, and 128 channels),
\item[] transposed-convolution layer with $3\times3$ filter, 3 channels and tanh activation.
\end{itemize}

\end{itemize}

For \our{}  architecture we used five feed-forward RELU layers with batch normalization and 512 neurons. The final layer was a~feed-forward layer with a~linear activation.

\section{Conclusions}

In this paper, we presented the new  \our{}. According to our knowledge, it is the first model that can be effectively trained using a~kernel distance in high dimensional data sets. It needs no adversarial training and does not suffer from ``mode collapse''. Consequently, the proposed approach might be situated between autoencoder and GAN models, while not inheriting their main weaknesses. 

Our experiments show that \our{} gives superior FID values for generated  samples. Another interesting feature is that it is a~natural generator of density-based interpolations in the latent space. Its ability to omit all low-density areas might give a robust tool for generating new samples with a smooth changing of features.

Finally, we want to stress that the \trick{} may be applied to any already trained autoencoder based model, to increase its generative capabilities.

\end{document}